\pdfoutput=1

\documentclass[11pt]{article}

\usepackage[preprint]{acl}

\usepackage{times}
\usepackage{latexsym}
\usepackage{subcaption}

\usepackage[T1]{fontenc}

\usepackage[utf8]{inputenc}

\usepackage{microtype}

\usepackage{inconsolata}

\usepackage{graphicx}

\usepackage{amsmath}
\usepackage{amssymb}

\title{Reducing hallucination in structured outputs via Retrieval-Augmented Generation}

\author{Patrice Béchard \\
  ServiceNow \\
  \texttt{patrice.bechard@servicenow.com} \\\And
 Orlando Marquez Ayala \\
  ServiceNow \\
  \texttt{orlando.marquez@servicenow.com} \\}

\begin{document}
\maketitle
\begin{abstract}
A current limitation of Generative AI (GenAI) is its propensity to hallucinate. While Large Language Models (LLM) have taken the world by storm, without eliminating or at least reducing hallucination, real-world GenAI systems will likely continue to face challenges in user adoption. In the process of deploying an enterprise application that produces workflows from natural language requirements, we devised a system leveraging Retrieval-Augmented Generation (RAG) to improve the quality of the structured output that represents such workflows. Thanks to our implementation of RAG, our proposed system significantly reduces hallucination and allows the generalization of our LLM  to out-of-domain settings. In addition, we show that using a small, well-trained retriever can reduce the size of the accompanying LLM at no loss in performance, thereby making deployments of LLM-based systems less resource-intensive. 
\end{abstract}

\section{Introduction}

With the advent of Large Language Models (LLMs), structured output tasks such as converting natural language to code or to SQL have become commercially viable. A similar application is translating a natural language requirement to a \textit{workflow}, a series of steps along with logic elements specifying their relationships. These workflows encapsulate processes that are executed automatically upon certain conditions, thereby increasing employee productivity. While enterprise systems offer such functionality to automate repetitive work and standardize processes, the barrier to entry is high, as building workflows requires specialized knowledge. Generative AI (GenAI) can lower this barrier since novice users can specify in natural language what they want their workflows to execute.

However, as with any GenAI application, using LLMs naively can produce \textbf{\textit{untrustworthy}} outputs. Such is the public concern for LLMs producing hallucinations that the Cambridge Dictionary chose \textit{hallucinate} as its Word of the Year in 2023 \cite{cambridge2023}. Retrieval-Augmented Generation (RAG) is a well-known method that can reduce hallucination and improve output quality, especially when generating the correct output requires access to external knowledge sources \cite{gao2024retrievalaugmented}.

\begin{figure}[t]
    \centering
    \includegraphics[width=\linewidth]{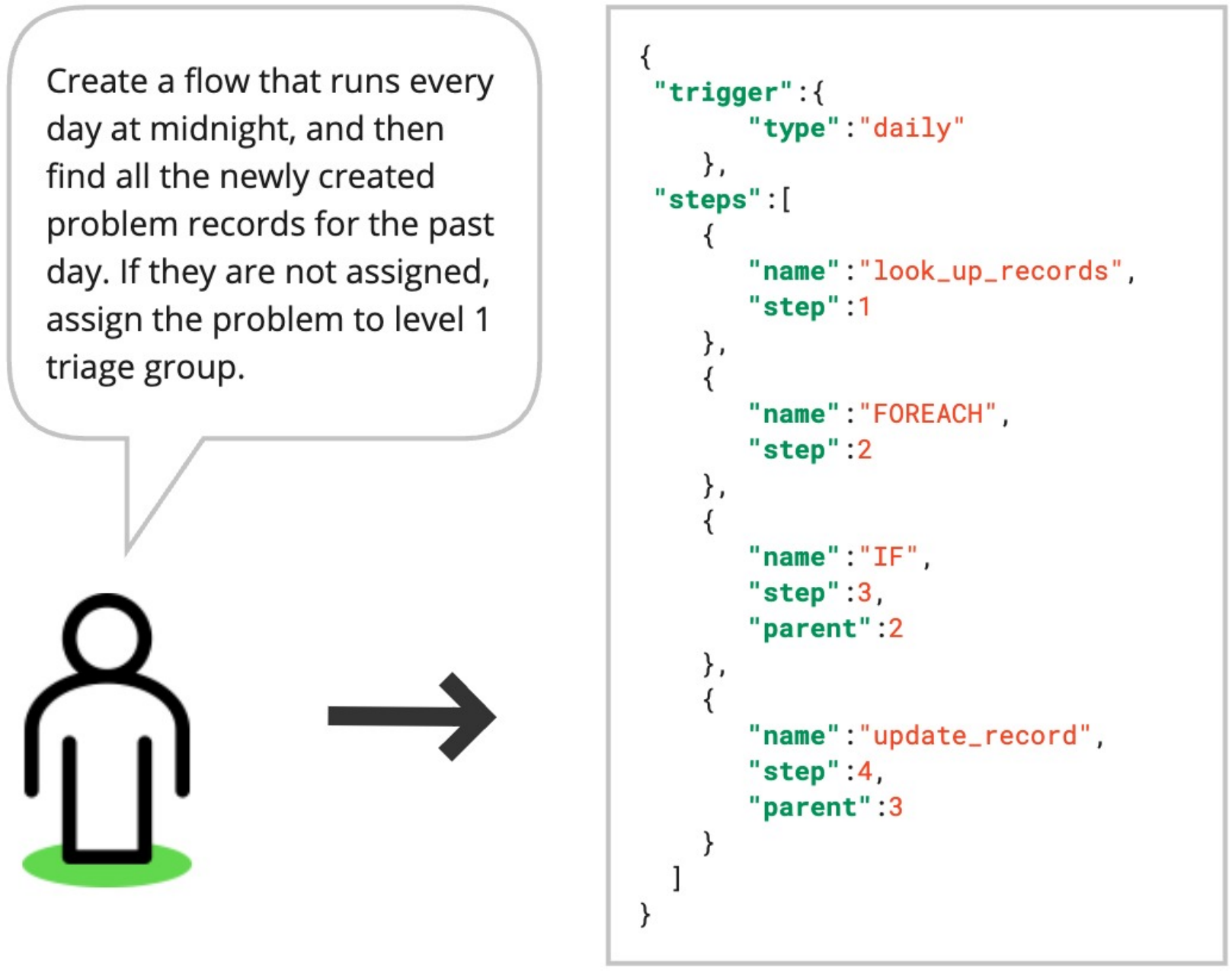}
    \caption{Sample structured output (JSON) to generate given a natural language requirement.}
    \label{fig:t2f_task_sample}
\vspace*{-0.5cm}
\end{figure}

In this work, we describe how, in the process of building a commercial application that converts natural language to workflows, we employ RAG to improve the trustworthiness of the output by reducing hallucination. Workflows are represented as JSON documents where each step is a JSON object. Figure \ref{fig:t2f_task_sample} shows an example of a text requirement and its associated JSON document. For simplicity, we include only the basic properties needed to identify a step along with properties indicating the relationship between steps. Besides the workflow steps, there may also be a \textit{trigger} step that determines when the workflow should start, and sometimes this trigger requires a database table name. Hallucination in this task means generating properties such as steps or tables that do not exist.

While fine-tuning a sufficiently large LLM can produce reasonably good workflows, the model may hallucinate, particularly if the natural language input is out-of-distribution. As the nature of enterprise users requires them to customize their applications, in this case by adding their own type of workflow steps, a commercial GenAI application needs to minimize the out-of-distribution mismatch. While one could fine-tune the LLM per enterprise, this may be prohibitively expensive due to the high infrastructure costs of fine-tuning LLMs. Another consideration when deploying LLMs is their footprint, making it preferable to deploy the smallest LLM that can perform the task.

Our contributions are the following:
\begin{itemize}
\itemsep-0.3em
\item We provide an application of RAG in workflow generation, a structured output task.
\item We show that using RAG reduces hallucination and improves results.
\item We demonstrate that RAG allows deploying a smaller LLM while using a very small retriever model, at no loss in performance.
\end{itemize}

\begin{figure*}[t]
    \centering
    \includegraphics[width=\textwidth]{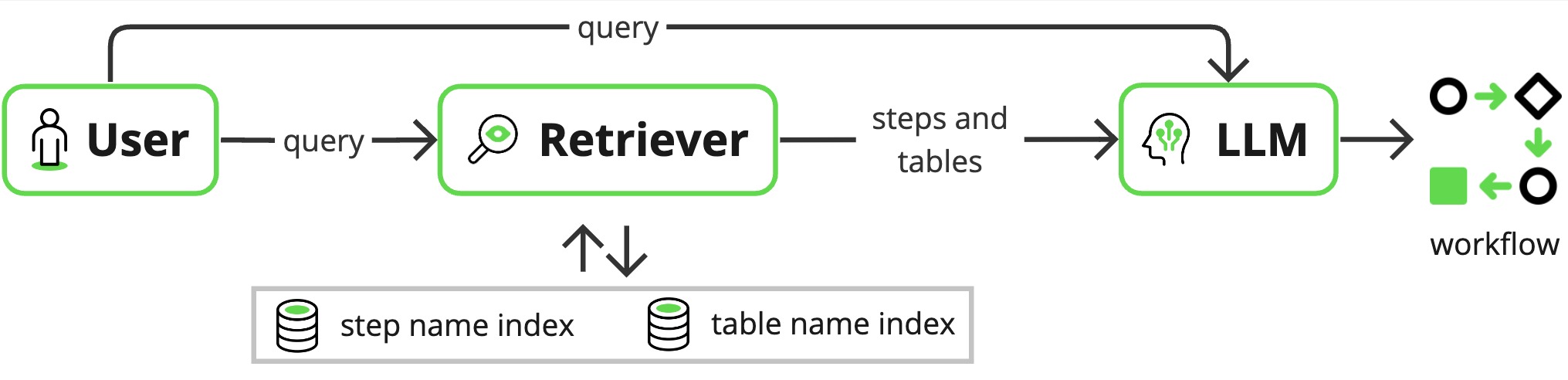}
    \caption{High-level architecture diagram showing how the user query is used by both the retriever and the LLM to generate the structured JSON output.}
    \label{fig:rag_diagram}
\vspace*{-0.2cm}
\end{figure*}

\section{Related Work}

\textbf{Retrieval-Augmented Generation} is a common approach to limit generation of false or outdated information in classical NLP tasks such as question answering and summarization \citep{lewis2020retrieval, izacard2021leveraging, shuster2021retrieval}. In the GenAI era, it refers to a process where relevant information from specific data sources is retrieved prior to generating text; the generation is then based on this retrieved information \cite{gao2024retrievalaugmented}. Our work differs from standard RAG as we apply it to a structured output task. Instead of retrieving facts, we retrieve JSON objects that could be part of the JSON output document. Providing plausible JSON objects to the LLM before generation increases the likelihood that the output JSON properties exist and that the generated JSON can be executed.

A crucial ingredient of RAG is the retriever since its output will be part of the LLM input. Compared to classical methods such as TF-IDF or BM25 that use lexical information, \textbf{Dense Retrieval} has been shown to be more effective as it maps the semantics to a multidimensional space where both queries and documents are represented \citep{reimers2019sentence, gao2021simcse, karpukhin2020dense, xiong2020approximate}. These retrievers are often used in open-domain question answering systems \citep{guu2020realm, lee2019latent}, where both queries and documents are unstructured data and thus share the same semantic space. In our case, the queries are unstructured (natural language) and the documents (JSON objects) are structured. Our retrieval training is similar to Structure Aware DeNse ReTrievAl (SANTA), which proposes a training method to align the semantics between code and text \cite{li-etal-2023-structure}.

Generating structured data falls within the realm of \textbf{Structured Output} tasks, which consist of generating a valid structured output from natural language, such as text-to-code, text-to-SQL \citep{zhong2017seq2sql, yu2018spider, wang2020rat} or if-then program synthesis \citep{quirk2015language, liu2016latent, dalal2020evaluating}. They are challenging as they not only require generating output that can be parsed, but also entities or field values that exist in a given lexicon; otherwise the resulting output cannot be interpreted or compiled. For simple database schemas or small lexicons, this extra information can be included in the prompt. However, in our task the available pool of steps that can be part of a workflow is potentially very large and customizable per deployment, thereby making in-context learning impractical.

With the arrival of LLMs, these tasks have become more accessible. In particular, \textbf{Code LLMs} enable developers to write code faster by providing instructions to the LLM to generate code snippets \citep{chen2021evaluating, nijkamp2022codegen, li2023starcoder, roziere2023code}. These models, trained on large datasets of source code \cite{kocetkov2022stack}, have acquired broad knowledge of many programming languages and have been shown to perform better at tasks that necessitate reasoning \cite{madaan2022language}. Since the JSON schema to represent workflows is domain-specific, we cannot use these models off-the-shelf. While fine-tuning them on a small dataset increases the quality of results, extra steps are required to reduce hallucination and support out-of-domain queries.

Lastly, an alternative and complementary technique to reduce hallucination with LLMs is \textbf{Guided Generation} using tools such as Outlines \cite{willard2023efficient}. A sufficiently expressive context-free grammar could ensure that the steps generated by the model exist, but it does not provide extra knowledge as to which steps the flow should include given the natural language query.

\section{Methodology}
Figure \ref{fig:rag_diagram} depicts the high-level architecture of our RAG system. During initialization, indices of steps and tables are created using the retriever. When a user submits a request, the retriever is called to suggest steps and tables. The suggestions are then appended to the user query to form the LLM prompt. The LLM is then called to generate the workflow in the JSON format via greedy decoding.

To build our system, we first train a retriever encoder to align natural language with JSON objects. We then train an LLM in a RAG fashion by including the retriever's output in its prompt.

\subsection{Retriever training}
We expect the LLM to learn to construct JSON documents including the relationship between workflow steps, given sufficient examples. The risk of hallucination comes mainly from the step names since there are tens of thousands of possible steps and every customer can add their own steps if the default set does not meet their needs. In addition, as some trigger steps require database table names as a property, these names can also be hallucinated. We therefore require the retriever to map natural language to existing step and database table names.

We choose to fine-tune a retriever model for two reasons: to improve the mapping between text and JSON objects, and to create a better representation of the domain of our application. While there exist a myriad of open-source sentence encoders \citep{reimers2019sentence, ni2022large}, they have been trained in a setting where both queries and documents are in the same natural language semantic space. But in our case, the query or workflow requirement is unstructured while the JSON objects are structured data. Consistent with the results reported by \citet{li-etal-2023-structure}, who search code snippets based on text, fine-tuning improves the retrieval results greatly. Similarly, fine-tuning a model using our domain-specific data allows the retriever to learn the nuances and technicalities of the text and JSON that are particular to our setting.

We use a siamese transformer encoder with mean pooling similar to \citet{reimers2019sentence} to encode both the user query and the step or table JSON object into fixed-length vectors. We include a normalization layer in our model so that the resulting embeddings have a norm of 1. We generate three embeddings $v_q\in \mathbb{R}^n$, $v_s\in \mathbb{R}^n$, $v_t\in \mathbb{R}^n$: 

\begin{equation}
    v_q = R(q) \qquad  v_s = R(s) \qquad v_t = R(t)
\end{equation}

where $q$, $s$, $t$ are the user query, step, and table respectively. Retriever $R$ can be decomposed as:

\begin{equation}
    R(q) = \textrm{Norm}(\textrm{MeanPool}(\textrm{Enc}(q)))
\end{equation}

The retriever model is trained on pairs of user queries and corresponding steps or tables. Since table names are used only in certain examples depending on the type of trigger, a query can be mapped to zero tables. For instance, the workflow in Figure \ref{fig:t2f_task_sample} has four steps, forming four positive training pairs, each pair consisting of the same query and one of the steps in the flow. As the \texttt{daily} trigger step does not need a table name, the query is mapped to an empty list of tables.

We also construct negative training pairs by sampling steps or tables that are not relevant to the user query. We experiment with three different negative sampling strategies: random, BM25-based, and ANCE-based \cite{xiong2020approximate}.

The retriever is trained using a contrastive loss \cite{hadsell2006dimensionality} to minimize the distance between positive pairs ($Y=1$) and negative pairs ($Y=0$). Given the cosine similarity between the query and step (or table) vectors, and cosine distance $D=1-\textrm{cossim}(v_q, v_s)$, we define contrastive loss $\mathcal{L}$ as:

\begin{equation}
    \mathcal{L} = \frac{1}{2}\left( YD^2 + (1-Y) \cdot \textrm{max}(0, \frac{1}{2} - D)^2 \right)
\end{equation}

During initialization, we build an index of steps and tables using FAISS \cite{douze2024faiss}. When a user submits a natural language query, we embed the incoming query using our retriever and use cosine similarity to retrieve the max $K$ steps and tables associated with this requirement.

\subsection{LLM training} 
Contrary to end-to-end RAG systems such as \citet{lewis2020retrieval}, we opted to train both the retriever and LLM separately, for simplicity. We use the trained retriever to augment our dataset with suggested step and table names for each example. We then proceed with standard LLM supervised fine-tuning. 

\begin{figure}[h]
    \centering
    \includegraphics[width=\linewidth]{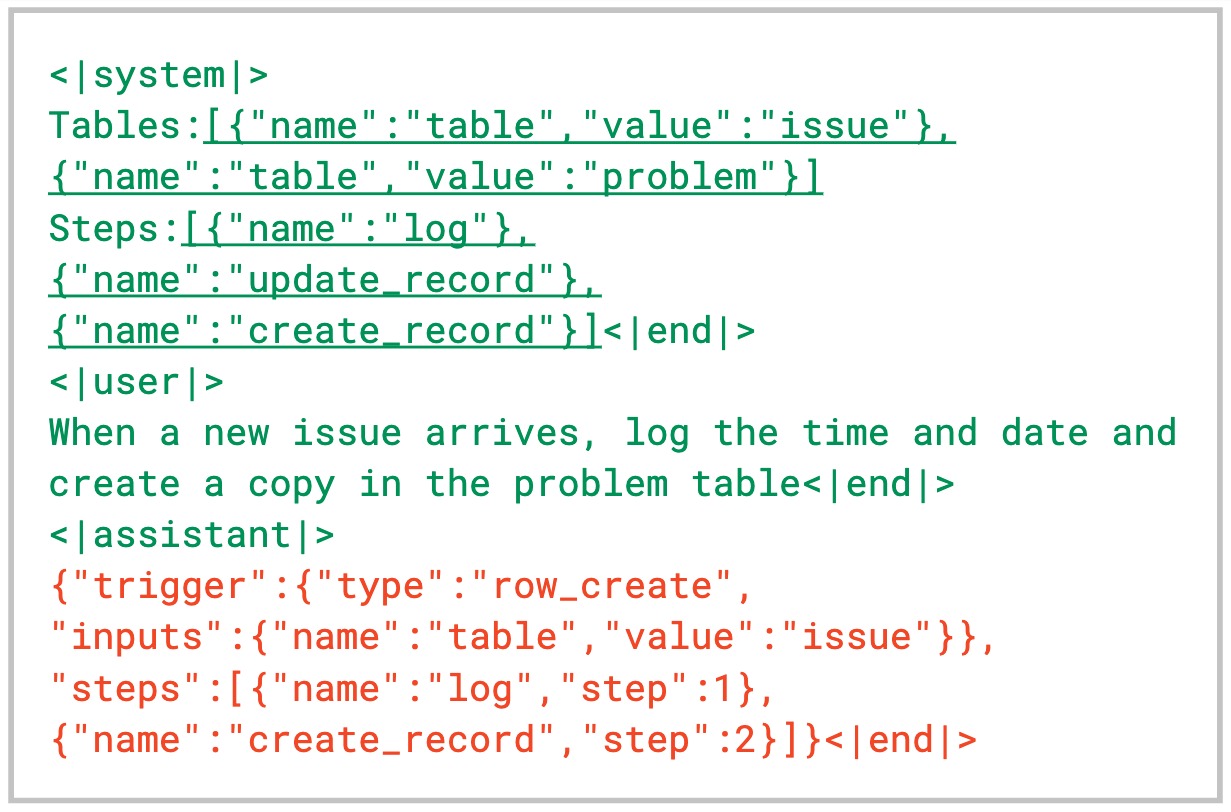}
    \caption{Training example, where the last four lines are the expected output (in red). The underlined text comes from the retriever's output.}
    \label{fig:input_output_example}
\end{figure}

By inserting the retriever's output in JSON format into the LLM input, we effectively make this structured output task easier as the LLM can copy the relevant JSON objects during generation. Figure \ref{fig:input_output_example} shows an example of a training example. Every line except the last four make up the LLM prompt. The suggested tables and steps come before the user query and are underlined in the figure. We exclude the most frequent steps from these suggestions as we expect the LLM to memorize them. Also, in every LLM training example, we assume the retriever has 100\% recall: the steps and table required to build the structured output are always in the suggestions, except for the most frequent steps.

As we are showing the LLM thousands of examples during training, we did not find it necessary to experiment with complicated or verbose prompts: we used a short and simple format, similar to Figure \ref{fig:input_output_example}, to reduce the number of input tokens while making it clear that this is a structured output task. As shown in section \ref{subsection-RAG-results}, this approach yielded good performance.
 
\section{Experiments}
As the task we are interested in is part of a commercial enterprise system, we had to devise our own datasets as well as evaluation metrics.

\subsection{Datasets}
From internal deployments of our enterprise platform, we extracted around 4,000 examples of deployed workflows and asked annotators to write natural language requirements for them. In addition, using deterministic rules, we created around 1,000 samples having simple and few steps in order to teach the model to handle input where the user is incrementally building their workflow. To have an unbiased estimate of the quality of results once the system is deployed, we asked expert users to simulate interacting with the system through a simple user interface where they typed their requirement. We used these interactions and the expected JSON documents to create an additional dataset split, named "Human Eval." Our final metrics are based on this split instead of the "Test" split, due to its higher quality and more realistic input. Table \ref{tab:indomain-stats} shows statistics for all of our in-domain splits. Not all samples require triggers, and a small subset require the model to generate tables.

\begin{table}[h]
\small
\centering
\begin{tabular}{lccc}
\hline
\textbf{Split}& \textbf{Size} & \textbf{\# Triggers} & \textbf{\# Tables} \\ \hline
Train  & 2867 & 823 & 556 \\
Dev  & 318 & 77  & 44 \\ 
Test & 798  & 247 & 163 \\
Human Eval & 157 & 99 & 60\\
\hline
\end{tabular}
\caption{Data statistics for in-domain training and evaluation.}
\label{tab:indomain-stats}
\vspace{-0.2cm}
\end{table}

A drawback of our data labeling approach is that these internal datasets are mostly in the IT domain, whereas our RAG system can be deployed in diverse domains such as HR and finance. Without assessing the quality of the system in out-of-distribution settings, we cannot be confident that the system will behave as expected. We therefore asked annotators to label five other splits, which come from other deployments of our enterprise platform. These are real workflows that have been created by real users.

Table \ref{tab:outdomain-stats} includes statistics for these out-of-domain splits. A measure of how different they are from our training data is the \% of steps that are not in the set of steps in the "Train" split. This discrepancy ranges from less than 10\% to more than 70\%, highlighting the need to use a retriever and to customize the indices per deployment.

\begin{table}[h]
\small
\begin{tabular}{lcccc}
\hline
\textbf{Split}& \textbf{Size} & \textbf{\# Triggers} & \textbf{\# Tables} & \textbf{\% Steps not} \\
& & & & \textbf{in Train}\\
\hline
OOD1 & 146 & 133 & 47 & 49\% \\
OOD2 & 162 & 111 & 21 & 76\% \\
OOD3 & 429 & 226 & 114 & 34\% \\
OOD4 & 42 & 25 & 11 & 33\% \\
OOD5 & 353 & 271 & 26 & 7\% \\
\hline
\end{tabular}
\caption{Data statistics for out-of-domain evaluation.}
\label{tab:outdomain-stats}
\vspace*{-0.1cm}
\end{table}

To train the retriever encoder, we create pair examples out of the 4,000 extracted and 1,000 deterministically generated samples, resulting in around 15,000 pairs in the step names dataset and 1,500 in the table names dataset. The quality of this encoder is evaluated on the "Human Eval" split described above.

\subsection{Metrics}
We evaluate the entire RAG system using three metrics, which can all range from 0 to 1:
\begin{itemize}
\itemsep-0.2em
\item \textbf{Trigger Exact Match (EM)} verifies whether the generated JSON trigger is exactly the same as the ground-truth, including the table name if this trigger requires it.
\item \textbf{Bag of Steps (BofS)} measures the overlap between the generated JSON steps and the ground-truth steps in an order-agnostic fashion, akin to a bag-of-words approach.
\item \textbf{Hallucinated Tables (HT)} and \textbf{Hallucinated Steps (HS)} measure the \% of generated tables/steps that do not exist per workflow, indicating that they were invented by the LLM. This is the only metric where lower is better.
\end{itemize}

To evaluate the retriever, we use \textbf{Recall@15} for steps and \textbf{Recall@10} for tables. That is, given a natural language requirement, we retrieve the top $K$ steps/tables from their respective indices and verify whether they cover the set of steps and the table, if required, included in the JSON document representing the workflow.

\subsection{Models}
As this is a production system, we have a trade-off between model size and performance for both the LLM and the retriever encoder.

We fine-tune models of different sizes to measure the impact of model size on the final metrics. As StarCoderBase \cite{li2023starcoder} has been pre-trained on JSON in addition to many programming languages and comes in different sizes, we fine-tune its 1B, 3B, 7B and 15.5B variants. Given our infrastructure constraints, we could deploy an LLM of at most 7B parameters. Thus we also fine-tune other pretrained LLMs of this size: CodeLlama-7B \cite{roziere2023code} and Mistral-7B-v0.1 \cite{jiang2023mistral}. All the LLMs were fine-tuned using the same datasets and hyperparameters.

We use all-mpnet-base-v2\footnote{https://huggingface.co/sentence-transformers/all-mpnet-base-v2} as the base retriever model. As it has only 110M parameters, it is suitable for deployment. We compare our fine-tuned model against different sizes of off-the-shelf GTR-T5 models \cite{ni2022large} to see whether larger encoders impact the performance.

Please see Appendix \ref{sec:app_train_details} for training details for both the LLM and the retriever encoder.

\section{Results}

\subsection{Retriever encoder}

Table \ref{tab:retriever} shows the results of retrieval on the "Human Eval" split for both steps and tables. Scaling the size of the off-the-shelf encoders, as we did with GTR-T5, does not yield significant improvements on both retrieval metrics. A similar observation was made by \citet{neelakantan2022text} for code retrieval. What was crucial to significantly improve the performance was fine-tuning the encoder.

\begin{table}[h]
\small
\centering
\begin{tabular}{lcc}
\hline
{ \textbf{Model (\# Params)}}&\textbf{Step} & \textbf{Table}\\
 & {\footnotesize \textbf{Recall@15}} & {\footnotesize \textbf{Recall@10}} \\
\hline
{gtr-t5-base (110M)} & 0.505 & 0.489 \\
{gtr-t5-large (355M)} & 0.575 & 0.511 \\
{gtr-t5-xl (1.24B)}& 0.579 & 0.489 \\
{gtr-t5-xxl (4.8B)} & 0.561 & 0.489 \\
\hline
{all-mpnet-base-v2 (110M)} & 0.425 & 0.170 \\
{\quad + Random} & 0.640 & 0.752\\
{\quad+ BM25} & 0.537 & 0.586\\ 
{\quad+ ANCE} & 0.556 & 0.699\\ 
{\quad+ All} & \textbf{0.743} & \textbf{0.766} \\\hline
\end{tabular}
\caption{Evaluation of different encoders on step and table retrieval. The last four rows represent encoders fine-tuned using different negative sampling strategies.}
\label{tab:retriever}
\vspace{-0.2cm}
\end{table}

Due to deployment considerations, we fine-tune the smallest encoders (110M parameters), and found that all-mpnet-base-v2 yielded the best performance after fine-tuning with all negative sampling strategies.

\begin{table*}[t]
\small
\centering
\begin{tabular}{lcccc}
\hline
& \textbf{Trigger} & \textbf{Bag of} & \textbf{Hallucinated} & \textbf{Hallucinated}\\
\textbf{Model} & \textbf{EM} & \textbf{Steps} & \textbf{Steps} & \textbf{Tables}\\
\hline
{\footnotesize \textbf{No Retriever}} \\
{StarCoderBase-1B} & 0.580 & 0.645 & 0.157 & 0.192\\
{StarCoderBase-3B} & 0.551 & 0.648 & 0.140 & 0.214\\
{StarCoderBase-7B} & 0.547 & \textbf{0.669} & 0.137 & 0.206\\
{StarCoderBase (15.5B)} & 0.632 & 0.662 & 0.160 & 0.194\\
\hline
{\footnotesize \textbf{With Retriever}} \\
{StarCoderBase-1B} & 0.591 & 0.619 & 0.072 & 0.044\\
{StarCoderBase-3B} & 0.615 & 0.641 & \textbf{0.017} & 0.030\\
{StarCoderBase-7B} & \textbf{0.664} & \textbf{0.672} & \textbf{0.019} & 0.042\\
{StarCoderBase (15.5B)} & \textbf{0.667} & \textbf{0.667} & 0.040 & \textbf{0.016}\\
{CodeLlama-7B} & 0.623 & 0.617 & 0.039 & 0.108 \\
{Mistral-7B-v0.1} & 0.596 & 0.617 & 0.049 & 0.045\\\hline
\end{tabular}
\caption{Performance of various model types and sizes on the "Human Eval" split. Lower is better for the hallucination metrics. Results within 0.005 of the best score are highlighted in \textbf{bold}.}
\label{tab:t2f_rag}
\vspace{-0.2cm}
\end{table*}

\subsection{Retrieval-Augmented Generation} \label{subsection-RAG-results}

Our main objective is to reduce hallucination while keeping the overall performance high given our infrastructure constraints. Table \ref{tab:t2f_rag} shows that without a retriever (only LLM fine-tuning), the \% of hallucinated steps and tables can be as high as 21\% on the "Human Eval" split. Using a retriever, this decreases to less than  7.5\% for steps and less than 4.5\% for tables with all StarCoderBase LLMs. All models produce valid JSON documents following the expected schema, thanks to fine-tuning.

Without a retriever, scaling the size of the StarCoderBase models improves the Bag of Steps and Trigger Exact Match metrics, albeit unevenly. Scaling also helps with RAG, but we observe more consistent improvements. This suggests that larger LLMs can better copy and paste retrieved steps and tables during generation.

The smallest RAG fine-tuned model (1B) hallucinates significantly more than its larger counterparts. Among the other three variants, the 7B version gives us the best trade-off, as the performance difference between 7B and 15.5B is marginal. Another observation is that the 3B version trained with RAG is competitive even with the 15.5B version without RAG on the Trigger EM and Bag of Steps metrics, while keeping hallucination low. This is a key lesson as we could deploy a 3B RAG fine-tuned model if we had more limited infrastructure. 

Lastly, we compare the RAG fine-tuned StarCoderBase-7B to fine-tuning more recent LLMs of the same size. Despite also fine-tuning them with RAG, CodeLlama-7B and Mistral-7B-v0.1 produce worse results across all metrics, even compared to the smaller StarCoderBase-3B. We suspect that pre-training on large amounts of natural language data may be detrimental to our task.

\subsection{OOD evaluation}

We want our approach to perform well on OOD scenarios without further fine-tuning the retriever or the LLM. Table \ref{tab:t2f_rag_customers} assesses the performance of our chosen RAG fine-tuned StarCoderBase-7B model on the five OOD splits described by Table \ref{tab:outdomain-stats}.

\begin{table}[h]
\small
\centering
\begin{tabular}{lccccc}
\hline
\textbf{Split} &  \textbf{Trigger EM} & \textbf{BofS} & \textbf{HS} & \textbf{HT} \\
\hline
{OOD1} & 0.662 & 0.619 & 0.063 & 0.051 \\
{OOD2} & 0.645 & 0.612 & 0.020 & 0.151 \\
{OOD3} & 0.562 & 0.743 & 0.014 & 0.033 \\
{OOD4} & 0.400 & 0.671 & 0.011 & 0.154 \\
{OOD5} & 0.774 & 0.770 & 0.005 & 0.063 \\
\hline
{Avg.} & 0.647 & 0.714 & 0.018 & 0.066\\
\hline
{No RAG Avg.} & 0.544 & 0.629 & 0.020 & 0.428\\
\hline
{Human Eval} & 0.664 & 0.672 & 0.019 & 0.042 \\
\hline
\end{tabular}
\caption{Performance of RAG fine-tuned StarCoderBase-7B on OOD splits.}
\label{tab:t2f_rag_customers}
\vspace{-0.2cm}
\end{table}

We observe that on average, thanks to the retriever, all the OOD metrics are similar to the in-domain results represented by the "Human Eval" split. We use a weighted average based on the number of samples per split.

To quantify the effect of suggesting step and table names, we evaluate the RAG fine-tuned StarCoderBase-7B model without suggestions in row "No RAG Avg.". All metrics worsen significantly while the "Hallucinated Steps" remains roughly the same. Upon inspection, we see that the RAG fine-tuned model has learned to be conservative in generating steps when it does not receive suggestions, relying only on steps that it has seen during training. On the other hand, the "Hallucinated Tables" metric is significantly worse as the model is more creative when it comes to tables. Please see Appendix B for supplementary detail.

\subsection{Error Analysis}
When investigating error patterns found in the generated workflows, we observe issues arising from failures both on the retriever and the LLM.  

For complex flows where steps that are used less frequently need to be retrieved, if a crucial component is not in the retriever's suggestions, it becomes difficult for the LLM to generate a valid workflow in line with the user query. To improve the retriever's recall, we can decompose the query into shorter texts to make the retrieval step more precise for each step. This would mean performing several retrieval calls, potentially one per step, instead of making one single retrieval call as we are doing now.

In some cases, the LLM did not produce the desired structure. This is more often seen when using steps that determine the logic of the workflow, such as \texttt{IF}, \texttt{TRY}, or \texttt{FOREACH}. These are important errors that can be addressed by synthetic data generation after analyzing which steps are being missed. For examples of perfect output and when the retriever and LLM fail, please refer to Appendix \ref{sec:app_error_analysis}.

\subsection{Impact on Engineering}

The obtained results led us to make several decisions that impacted the scalability and modularity of the system. Since the best overall performance was given by a 7B-parameter model, we could have a larger batch size for incoming user requests, thereby increasing the system throughput given a single GPU. This implies a trade-off in latency as larger queries (in number of tokens) result in larger number of generated tokens, sometimes causing large queries to become a bottleneck if they are included in a batch with many shorter queries. Our stress tests and user research reveal that the current system overall response time is acceptable.

Obtaining good results after fine-tuning a very small encoder for the retriever (110M parameters), allowed us to deploy it on the same GPU with negligible effect on the larger LLM. But we could even deploy the retriever on CPU due to its small size. A benefit of not performing joint training between the retriever and the LLM is that the retriever can be reused for other use cases involving similar data sources. Moreover, decoupling them allows clearer separation of concerns and independent optimization by separate team members. Nevertheless, for scientific purposes, it is still worthwhile to experiment with joint training.

We have several ideas to reduce the system response time: changing the structured output format from JSON to YAML to reduce the number of tokens, leveraging speculative decoding \cite{leviathan2023fast, chen2023accelerating, gante2023assisted}, and streaming one step at a time back to the user instead of the entire generated workflow.

\section{Conclusion}

We propose an approach to deploy a Retrieval-Augmented LLM to reduce hallucination and allow generalization in a structured output task. Reducing hallucination is a sine qua non for users to adopt real-world GenAI systems. We show that RAG allows deploying a system in limited-resource settings as a very small retriever can be coupled with a small LLM. Future work includes improving the synergy between the retriever and the LLM, through joint training or a model architecture that allows them to work better together.

\section*{Ethical Considerations}
While our work proposes an approach to reduce hallucination in structure output tasks, we do not claim that the risk of harm due to hallucination is eliminated. Our deployed system includes a layer of post-processing to clearly indicate to users the generated steps that do not exist and urge them to fix the output before continuing their work.

\section*{Acknowledgements}
We thank our ServiceNow colleagues who worked hard in building the aforementioned system, from project managers to quality engineers. We also thank the several colleagues who reviewed an earlier version of this paper: Lindsay Brin, Hessam Amini, Erfan Hosseini, and Gabrielle Gauthier-Melançon, as well as the NAACL reviewers, for their valuable feedback.

\bibliography{custom}

\begin{thebibliography}{36}
\expandafter\ifx\csname natexlab\endcsname\relax\def\natexlab#1{#1}\fi

\bibitem[{Cambridge(2023)}]{cambridge2023}
Cambridge. 2023.
\newblock Why hallucinate?
\newblock \url{https://dictionary.cambridge.org/editorial/woty}.

\bibitem[{Chen et~al.(2023)Chen, Borgeaud, Irving, Lespiau, Sifre, and Jumper}]{chen2023accelerating}
Charlie Chen, Sebastian Borgeaud, Geoffrey Irving, Jean-Baptiste Lespiau, Laurent Sifre, and John Jumper. 2023.
\newblock \href {http://arxiv.org/abs/2302.01318} {Accelerating large language model decoding with speculative sampling}.

\bibitem[{Chen et~al.(2021)Chen, Tworek, Jun, Yuan, Pinto, Kaplan, Edwards, Burda, Joseph, Brockman et~al.}]{chen2021evaluating}
Mark Chen, Jerry Tworek, Heewoo Jun, Qiming Yuan, Henrique Ponde de~Oliveira Pinto, Jared Kaplan, Harri Edwards, Yuri Burda, Nicholas Joseph, Greg Brockman, et~al. 2021.
\newblock Evaluating large language models trained on code.
\newblock \emph{arXiv preprint arXiv:2107.03374}.

\bibitem[{Dalal and Galbraith(2020)}]{dalal2020evaluating}
Dhairya Dalal and Byron~V Galbraith. 2020.
\newblock Evaluating sequence-to-sequence learning models for if-then program synthesis.
\newblock \emph{arXiv preprint arXiv:2002.03485}.

\bibitem[{Dao et~al.(2022)Dao, Fu, Ermon, Rudra, and R{\'e}}]{dao2022flashattention}
Tri Dao, Dan Fu, Stefano Ermon, Atri Rudra, and Christopher R{\'e}. 2022.
\newblock Flashattention: Fast and memory-efficient exact attention with io-awareness.
\newblock \emph{Advances in Neural Information Processing Systems}, 35:16344--16359.

\bibitem[{Douze et~al.(2024)Douze, Guzhva, Deng, Johnson, Szilvasy, Mazaré, Lomeli, Hosseini, and Jégou}]{douze2024faiss}
Matthijs Douze, Alexandr Guzhva, Chengqi Deng, Jeff Johnson, Gergely Szilvasy, Pierre-Emmanuel Mazaré, Maria Lomeli, Lucas Hosseini, and Hervé Jégou. 2024.
\newblock \href {http://arxiv.org/abs/2401.08281} {The faiss library}.

\bibitem[{Gao et~al.(2021)Gao, Yao, and Chen}]{gao2021simcse}
Tianyu Gao, Xingcheng Yao, and Danqi Chen. 2021.
\newblock Simcse: Simple contrastive learning of sentence embeddings.
\newblock In \emph{2021 Conference on Empirical Methods in Natural Language Processing, EMNLP 2021}, pages 6894--6910. Association for Computational Linguistics (ACL).

\bibitem[{Gao et~al.(2024)Gao, Xiong, Gao, Jia, Pan, Bi, Dai, Sun, Guo, Wang, and Wang}]{gao2024retrievalaugmented}
Yunfan Gao, Yun Xiong, Xinyu Gao, Kangxiang Jia, Jinliu Pan, Yuxi Bi, Yi~Dai, Jiawei Sun, Qianyu Guo, Meng Wang, and Haofen Wang. 2024.
\newblock \href {http://arxiv.org/abs/2312.10997} {Retrieval-augmented generation for large language models: A survey}.

\bibitem[{Guu et~al.(2020)Guu, Lee, Tung, Pasupat, and Chang}]{guu2020realm}
Kelvin Guu, Kenton Lee, Zora Tung, Panupong Pasupat, and Ming-Wei Chang. 2020.
\newblock Realm: retrieval-augmented language model pre-training.
\newblock In \emph{Proceedings of the 37th International Conference on Machine Learning}, pages 3929--3938.

\bibitem[{Hadsell et~al.(2006)Hadsell, Chopra, and LeCun}]{hadsell2006dimensionality}
Raia Hadsell, Sumit Chopra, and Yann LeCun. 2006.
\newblock Dimensionality reduction by learning an invariant mapping.
\newblock In \emph{2006 IEEE Computer Society Conference on Computer Vision and Pattern Recognition, CVPR 2006}, pages 1735--1742.

\bibitem[{Hu et~al.(2021)Hu, Wallis, Allen-Zhu, Li, Wang, Wang, Chen et~al.}]{hu2021lora}
Edward~J Hu, Phillip Wallis, Zeyuan Allen-Zhu, Yuanzhi Li, Shean Wang, Lu~Wang, Weizhu Chen, et~al. 2021.
\newblock Lora: Low-rank adaptation of large language models.
\newblock In \emph{International Conference on Learning Representations}.

\bibitem[{Izacard and Grave(2021)}]{izacard2021leveraging}
Gautier Izacard and Edouard Grave. 2021.
\newblock Leveraging passage retrieval with generative models for open domain question answering.
\newblock In \emph{EACL 2021-16th Conference of the European Chapter of the Association for Computational Linguistics}, pages 874--880. Association for Computational Linguistics.

\bibitem[{Jiang et~al.(2023)Jiang, Sablayrolles, Mensch, Bamford, Chaplot, Casas, Bressand, Lengyel, Lample, Saulnier et~al.}]{jiang2023mistral}
Albert~Q Jiang, Alexandre Sablayrolles, Arthur Mensch, Chris Bamford, Devendra~Singh Chaplot, Diego de~las Casas, Florian Bressand, Gianna Lengyel, Guillaume Lample, Lucile Saulnier, et~al. 2023.
\newblock Mistral 7b.
\newblock \emph{arXiv preprint arXiv:2310.06825}.

\bibitem[{{Joao Gante}(2023)}]{gante2023assisted}
{Joao Gante}. 2023.
\newblock \href {https://doi.org/10.57967/hf/0638} {Assisted generation: a new direction toward low-latency text generation}.

\bibitem[{Karpukhin et~al.(2020)Karpukhin, Oguz, Min, Lewis, Wu, Edunov, Chen, and Yih}]{karpukhin2020dense}
Vladimir Karpukhin, Barlas Oguz, Sewon Min, Patrick Lewis, Ledell Wu, Sergey Edunov, Danqi Chen, and Wen-tau Yih. 2020.
\newblock Dense passage retrieval for open-domain question answering.
\newblock In \emph{Proceedings of the 2020 Conference on Empirical Methods in Natural Language Processing (EMNLP)}. Association for Computational Linguistics.

\bibitem[{Kocetkov et~al.(2022)Kocetkov, Li, Jia, Mou, Jernite, Mitchell, Ferrandis, Hughes, Wolf, Bahdanau et~al.}]{kocetkov2022stack}
Denis Kocetkov, Raymond Li, LI~Jia, Chenghao Mou, Yacine Jernite, Margaret Mitchell, Carlos~Mu{\~n}oz Ferrandis, Sean Hughes, Thomas Wolf, Dzmitry Bahdanau, et~al. 2022.
\newblock The stack: 3 tb of permissively licensed source code.
\newblock \emph{Transactions on Machine Learning Research}.

\bibitem[{Lee et~al.(2019)Lee, Chang, and Toutanova}]{lee2019latent}
Kenton Lee, Ming-Wei Chang, and Kristina Toutanova. 2019.
\newblock Latent retrieval for weakly supervised open domain question answering.
\newblock In \emph{Proceedings of the 57th Annual Meeting of the Association for Computational Linguistics}, pages 6086--6096.

\bibitem[{Leviathan et~al.(2023)Leviathan, Kalman, and Matias}]{leviathan2023fast}
Yaniv Leviathan, Matan Kalman, and Yossi Matias. 2023.
\newblock \href {http://arxiv.org/abs/2211.17192} {Fast inference from transformers via speculative decoding}.

\bibitem[{Lewis et~al.(2020)Lewis, Perez, Piktus, Petroni, Karpukhin, Goyal, K{\"u}ttler, Lewis, Yih, Rockt{\"a}schel et~al.}]{lewis2020retrieval}
Patrick Lewis, Ethan Perez, Aleksandra Piktus, Fabio Petroni, Vladimir Karpukhin, Naman Goyal, Heinrich K{\"u}ttler, Mike Lewis, Wen-tau Yih, Tim Rockt{\"a}schel, et~al. 2020.
\newblock Retrieval-augmented generation for knowledge-intensive nlp tasks.
\newblock In \emph{Proceedings of the 34th International Conference on Neural Information Processing Systems}, pages 9459--9474.

\bibitem[{Li et~al.(2023{\natexlab{a}})Li, Allal, Zi, Muennighoff, Kocetkov, Mou, Marone, Akiki, Li, Chim et~al.}]{li2023starcoder}
Raymond Li, Loubna~Ben Allal, Yangtian Zi, Niklas Muennighoff, Denis Kocetkov, Chenghao Mou, Marc Marone, Christopher Akiki, Jia Li, Jenny Chim, et~al. 2023{\natexlab{a}}.
\newblock Starcoder: may the source be with you!
\newblock \emph{arXiv preprint arXiv:2305.06161}.

\bibitem[{Li et~al.(2023{\natexlab{b}})Li, Liu, Xiong, Yu, Gu, Liu, and Yu}]{li-etal-2023-structure}
Xinze Li, Zhenghao Liu, Chenyan Xiong, Shi Yu, Yu~Gu, Zhiyuan Liu, and Ge~Yu. 2023{\natexlab{b}}.
\newblock \href {https://doi.org/10.18653/v1/2023.findings-acl.734} {Structure-aware language model pretraining improves dense retrieval on structured data}.
\newblock In \emph{Findings of the Association for Computational Linguistics: ACL 2023}, pages 11560--11574, Toronto, Canada. Association for Computational Linguistics.

\bibitem[{Liu et~al.(2016)Liu, Chen, Shin, Chen, and Song}]{liu2016latent}
Chang Liu, Xinyun Chen, Eui~Chul Shin, Mingcheng Chen, and Dawn Song. 2016.
\newblock Latent attention for if-then program synthesis.
\newblock \emph{Advances in Neural Information Processing Systems}, 29.

\bibitem[{Loshchilov and Hutter(2018)}]{loshchilov2018decoupled}
Ilya Loshchilov and Frank Hutter. 2018.
\newblock Decoupled weight decay regularization.
\newblock In \emph{International Conference on Learning Representations}.

\bibitem[{Madaan et~al.(2022)Madaan, Zhou, Alon, Yang, and Neubig}]{madaan2022language}
Aman Madaan, Shuyan Zhou, Uri Alon, Yiming Yang, and Graham Neubig. 2022.
\newblock Language models of code are few-shot commonsense learners.
\newblock In \emph{Proceedings of the 2022 Conference on Empirical Methods in Natural Language Processing}, pages 1384--1403.

\bibitem[{Neelakantan et~al.(2022)Neelakantan, Xu, Puri, Radford, Han, Tworek, Yuan, Tezak, Kim, Hallacy et~al.}]{neelakantan2022text}
Arvind Neelakantan, Tao Xu, Raul Puri, Alec Radford, Jesse~Michael Han, Jerry Tworek, Qiming Yuan, Nikolas Tezak, Jong~Wook Kim, Chris Hallacy, et~al. 2022.
\newblock Text and code embeddings by contrastive pre-training.
\newblock \emph{arXiv preprint arXiv:2201.10005}.

\bibitem[{Ni et~al.(2022)Ni, Qu, Lu, Dai, Abrego, Ma, Zhao, Luan, Hall, Chang et~al.}]{ni2022large}
Jianmo Ni, Chen Qu, Jing Lu, Zhuyun Dai, Gustavo~Hernandez Abrego, Ji~Ma, Vincent Zhao, Yi~Luan, Keith Hall, Ming-Wei Chang, et~al. 2022.
\newblock Large dual encoders are generalizable retrievers.
\newblock In \emph{Proceedings of the 2022 Conference on Empirical Methods in Natural Language Processing}, pages 9844--9855.

\bibitem[{Nijkamp et~al.(2022)Nijkamp, Pang, Hayashi, Tu, Wang, Zhou, Savarese, and Xiong}]{nijkamp2022codegen}
Erik Nijkamp, Bo~Pang, Hiroaki Hayashi, Lifu Tu, Huan Wang, Yingbo Zhou, Silvio Savarese, and Caiming Xiong. 2022.
\newblock Codegen: An open large language model for code with multi-turn program synthesis.
\newblock In \emph{The Eleventh International Conference on Learning Representations}.

\bibitem[{Quirk et~al.(2015)Quirk, Mooney, and Galley}]{quirk2015language}
Chris Quirk, Raymond Mooney, and Michel Galley. 2015.
\newblock Language to code: Learning semantic parsers for if-this-then-that recipes.
\newblock In \emph{Proceedings of the 53rd Annual Meeting of the Association for Computational Linguistics and the 7th International Joint Conference on Natural Language Processing (Volume 1: Long Papers)}, pages 878--888.

\bibitem[{Reimers and Gurevych(2019)}]{reimers2019sentence}
Nils Reimers and Iryna Gurevych. 2019.
\newblock Sentence-bert: Sentence embeddings using siamese bert-networks.
\newblock In \emph{Proceedings of the 2019 Conference on Empirical Methods in Natural Language Processing and the 9th International Joint Conference on Natural Language Processing (EMNLP-IJCNLP)}. Association for Computational Linguistics.

\bibitem[{Roziere et~al.(2023)Roziere, Gehring, Gloeckle, Sootla, Gat, Tan, Adi, Liu, Remez, Rapin et~al.}]{roziere2023code}
Baptiste Roziere, Jonas Gehring, Fabian Gloeckle, Sten Sootla, Itai Gat, Xiaoqing~Ellen Tan, Yossi Adi, Jingyu Liu, Tal Remez, J{\'e}r{\'e}my Rapin, et~al. 2023.
\newblock Code llama: Open foundation models for code.
\newblock \emph{arXiv preprint arXiv:2308.12950}.

\bibitem[{Shuster et~al.(2021)Shuster, Poff, Chen, Kiela, and Weston}]{shuster2021retrieval}
Kurt Shuster, Spencer Poff, Moya Chen, Douwe Kiela, and Jason Weston. 2021.
\newblock Retrieval augmentation reduces hallucination in conversation.
\newblock In \emph{Findings of the Association for Computational Linguistics: EMNLP 2021}, pages 3784--3803.

\bibitem[{Wang et~al.(2020)Wang, Shin, Liu, Polozov, and Richardson}]{wang2020rat}
Bailin Wang, Richard Shin, Xiaodong Liu, Oleksandr Polozov, and Matthew Richardson. 2020.
\newblock Rat-sql: Relation-aware schema encoding and linking for text-to-sql parsers.
\newblock In \emph{Proceedings of the 58th Annual Meeting of the Association for Computational Linguistics}, pages 7567--7578.

\bibitem[{Willard and Louf(2023)}]{willard2023efficient}
Brandon~T Willard and R{\'e}mi Louf. 2023.
\newblock Efficient guided generation for llms.
\newblock \emph{arXiv preprint arXiv:2307.09702}.

\bibitem[{Xiong et~al.(2020)Xiong, Xiong, Li, Tang, Liu, Bennett, Ahmed, and Overwijk}]{xiong2020approximate}
Lee Xiong, Chenyan Xiong, Ye~Li, Kwok-Fung Tang, Jialin Liu, Paul~N Bennett, Junaid Ahmed, and Arnold Overwijk. 2020.
\newblock Approximate nearest neighbor negative contrastive learning for dense text retrieval.
\newblock In \emph{International Conference on Learning Representations}.

\bibitem[{Yu et~al.(2018)Yu, Zhang, Yang, Yasunaga, Wang, Li, Ma, Li, Yao, Roman et~al.}]{yu2018spider}
Tao Yu, Rui Zhang, Kai Yang, Michihiro Yasunaga, Dongxu Wang, Zifan Li, James Ma, Irene Li, Qingning Yao, Shanelle Roman, et~al. 2018.
\newblock Spider: A large-scale human-labeled dataset for complex and cross-domain semantic parsing and text-to-sql task.
\newblock In \emph{Proceedings of the 2018 Conference on Empirical Methods in Natural Language Processing}, pages 3911--3921.

\bibitem[{Zhong et~al.(2017)Zhong, Xiong, and Socher}]{zhong2017seq2sql}
Victor Zhong, Caiming Xiong, and Richard Socher. 2017.
\newblock Seq2sql: Generating structured queries from natural language using reinforcement learning.
\newblock \emph{arXiv preprint arXiv:1709.00103}.

\end{thebibliography}
\clearpage

\appendix

\section{Training details for LLM and retriever}
\label{sec:app_train_details}

All LLMs were fine-tuned using the same set of hyperparameters. We use the AdamW optimizer with a learning rate of $5e-4$, $\beta_1=0.9$, $\beta_2=0.999$ and weight decay of $0.01$. Models were trained for 5,000 steps with a cosine learning rate scheduler with 100 warmup steps. We use an effective batch size of 32 for all models, using gradient accumulation when the batch size would not fit on a single GPU. We trained all models using LoRA \cite{hu2021lora} with $r=16$, $\alpha=16$ and a dropout rate of $0.05$. All models were trained with flash-attention \cite{dao2022flashattention} on a single A100 80GB GPU.

We fine-tuned the retriever model using the SentenceTransformers framework \cite{reimers2019sentence}. We use the AdamW optimizer \cite{loshchilov2018decoupled} and a learning rate of $2e-5$. We use a batch size of 128 and train the model for 10 epochs.

\section{Differences in generation with and without suggestions}
To understand the impact of suggesting step and table names during generation, for each OOD split, we inspect the \% of unique steps and \% of unique table names that are hallucinated with and without suggestions.

Table \ref{tab:steps_hallucinated} shows that without suggestions, the RAG fine-tuned StarCoderBase-7B tends to generate significantly fewer unique step names. Receiving suggestions allows the model to copy the suggestions, thereby increasing the diversity of what it generates. In addition, without suggestions a greater percentage of the unique step names it generates are invented.

\begin{table}[h]
\small
\centering
\begin{tabular}{lcccc}
\hline
 & \multicolumn{2}{c}{\textbf{No suggestions}} & \multicolumn{2}{c}{\textbf{With suggestions}}\\
 \hline
\textbf{Split} & \# unique & \% H & \# unique & \% H\\
 & steps & & steps\\
 \hline
\textbf{OOD1} & 52 & 40\% & 100 & 13\% \\
\textbf{OOD2} & 38 & 34\% & 96 & 13\% \\
\textbf{OOD3} & 122 & 37\% & 269 & 9\% \\
\textbf{OOD4} & 20 & 5\% & 32 & 9\% \\
\textbf{OOD5} & 88 & 17\% & 151 & 3\% \\
\hline
\end{tabular}
\caption{Statistics of generated step names in terms of uniqueness and hallucination. H refers to unique hallucinated step names.}
\label{tab:steps_hallucinated}
\end{table}

We also see that even with suggestions, there is still an important gap in the percentage of unique step names that are hallucinated, as in some splits more than 10\% of unique steps are invented. While the overall hallucination rate is less than 2\%, as shown in Table \ref{tab:t2f_rag_customers}, there are cases where the retriever does not suggest what is expected or the LLM does not take into account the suggestions.

\begin{table}[h]
\small
\centering
\begin{tabular}{lcccc}
\hline
 & \multicolumn{2}{c}{\textbf{No suggestions}} & \multicolumn{2}{c}{\textbf{With suggestions}}\\
 \hline
\textbf{Split} & \# unique & \% H & \# unique & \% H\\
 & tables & & tables\\
 \hline
\textbf{OOD1} & 40 & 70\% & 22 & 14\% \\
\textbf{OOD2} & 31 & 71\% & 19 & 21\% \\
\textbf{OOD3} & 61 & 64\% & 44 & 9\% \\
\textbf{OOD4} & 11 & 54\% & 9 & 22\% \\
\textbf{OOD5} & 38 & 68\% & 29 & 17\% \\
\hline
\end{tabular}
\caption{Statistics of generated table names in terms of uniqueness and hallucination. H refers to unique hallucinated table names}
\label{tab:tables_hallucinated}
\end{table}

When it comes to table names, there are similar and different observations, as shown in Table \ref{tab:tables_hallucinated}. As in the case of step names, without suggestions a greater percentage of unique table names are invented. However, when provided with suggestions, the model is more conservative as it generates fewer unique table names. This may be an artifact of the data, where there is less diversity of tables used compared to step names.

\begin{figure*}[t]
\vspace{-10cm}
\centering
     \begin{subfigure}[b]{.3\textwidth}
         \centering
         \includegraphics[width=\textwidth]{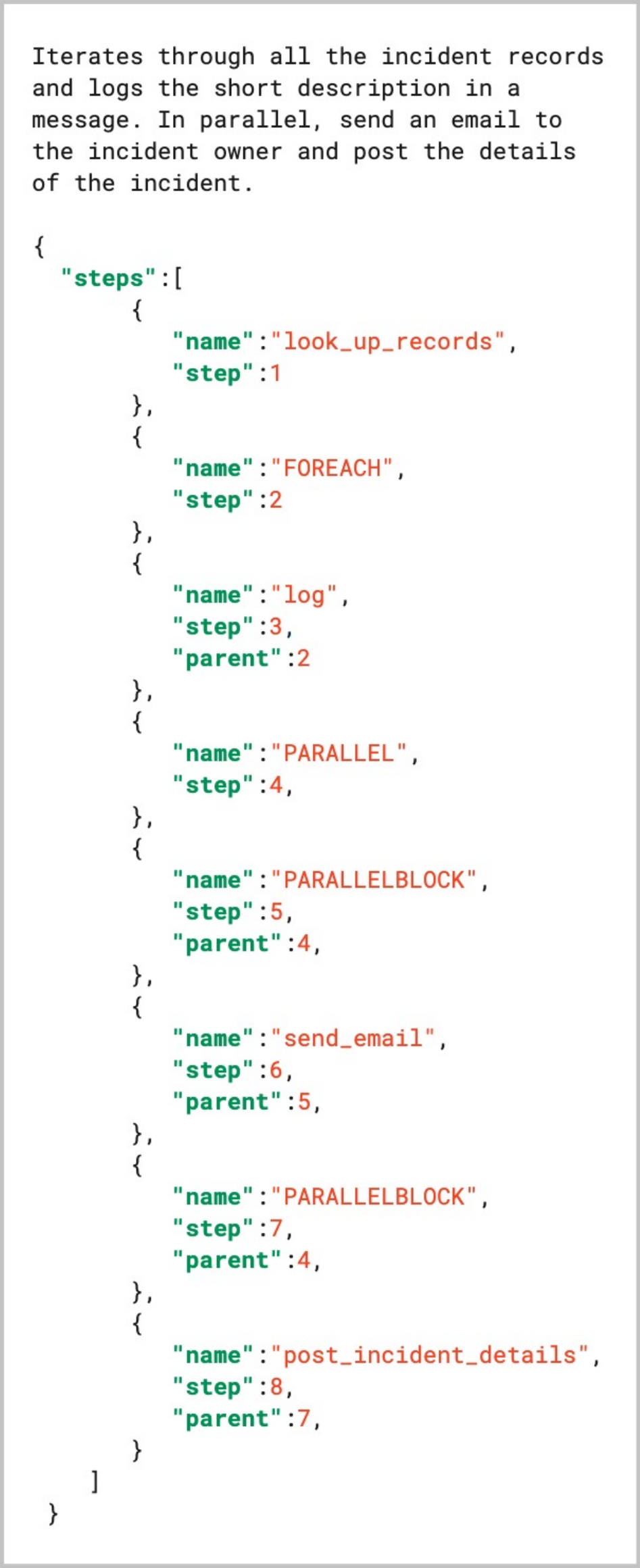}
         \caption{Perfect output}
         \label{fig:y equals x}
     \end{subfigure}
     \hfill
     \begin{subfigure}[b]{.3\textwidth}
         \centering
         \includegraphics[width=\textwidth]{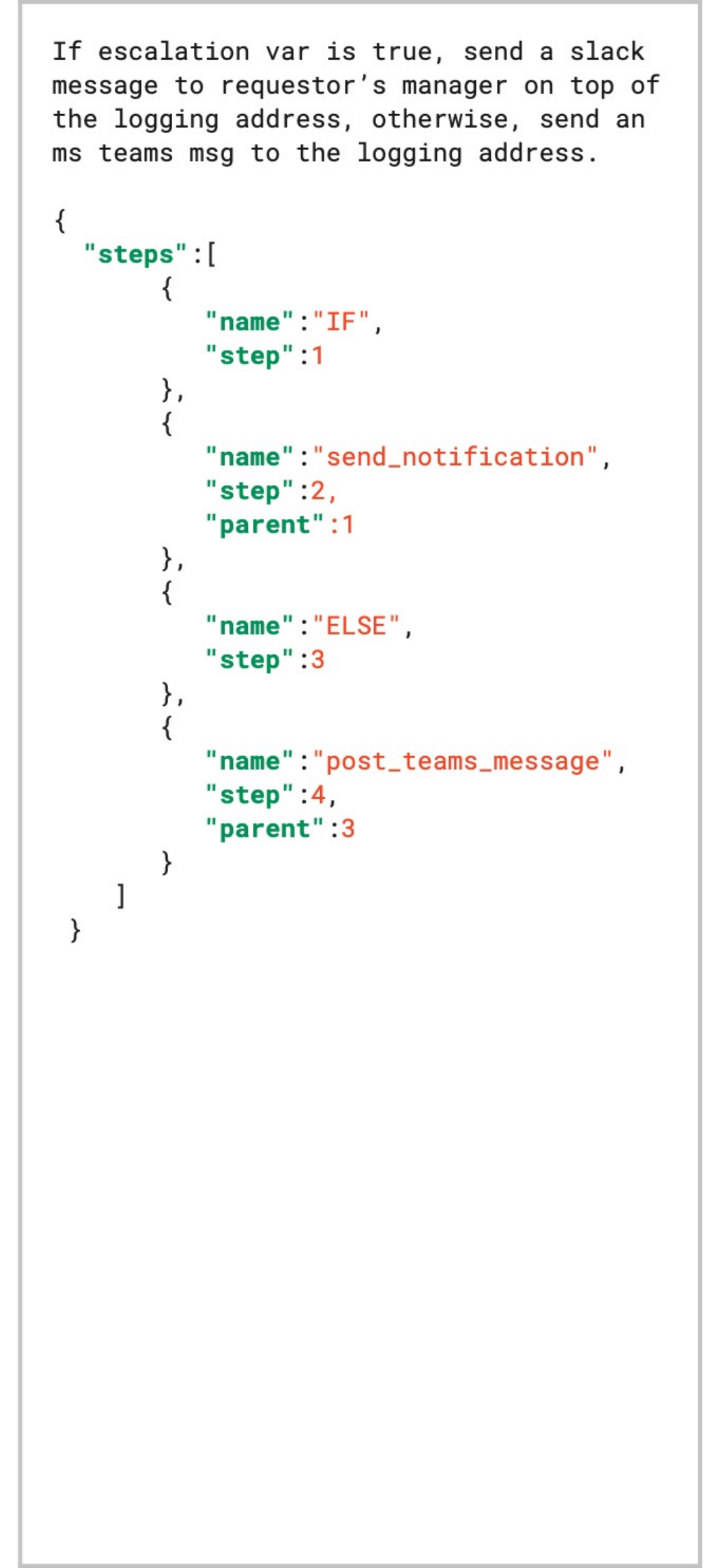}
         \caption{Retrieval error}
         \label{fig:three sin x}
     \end{subfigure}
     \hfill
     \begin{subfigure}[b]{.3\textwidth}
         \centering
         \includegraphics[width=\textwidth]{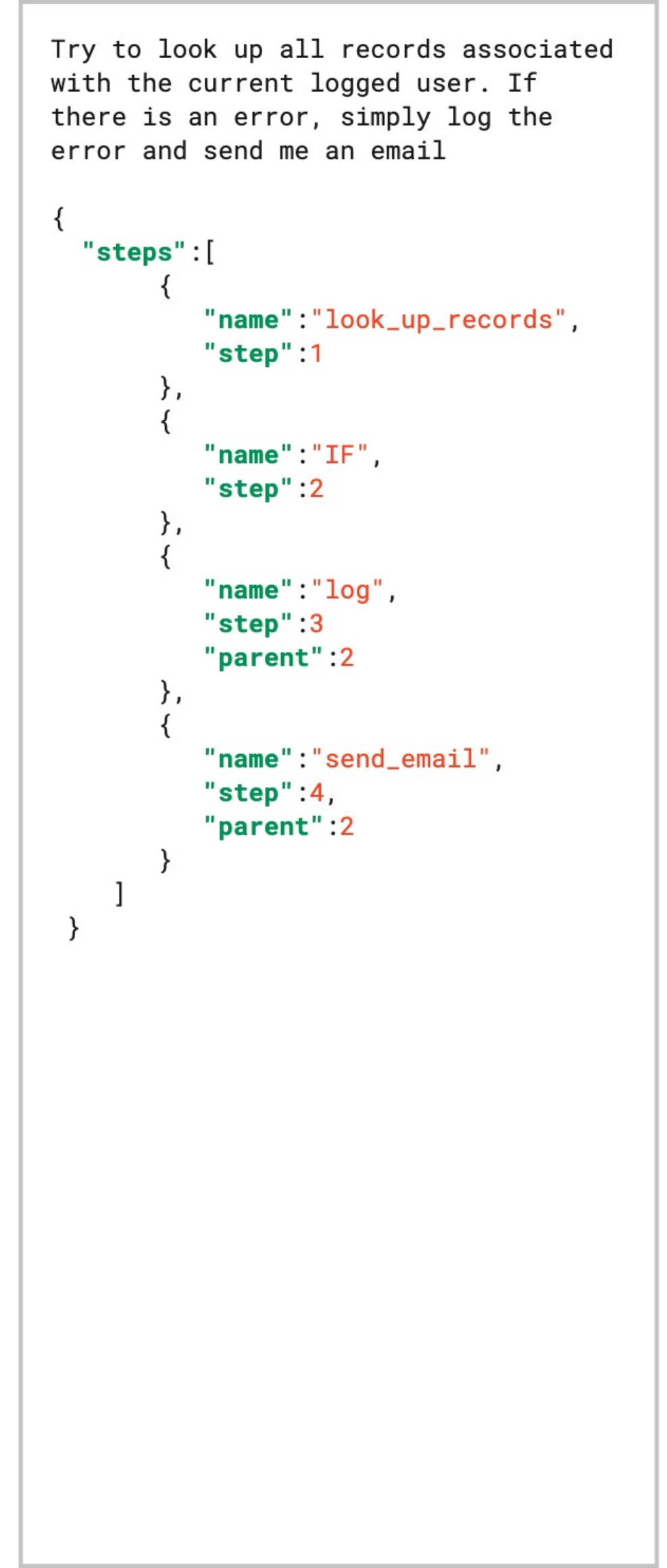}
         \caption{LLM error}
         \label{fig:five over x}
     \end{subfigure}
        \caption{Examples where both the retriever and the LLM worked perfectly and where each of them failed: (a) All expected step names were suggested and used by the LLM. (b) The retriever did not suggest the step \texttt{send\_slack\_message} and therefore the LLM used the common step \texttt{send\_notification} instead. (c) The LLM should have used the \texttt{TRY} step as the parent to all the steps, but it did not fully understand the user query.}
        \label{fig:error_analysis}
\end{figure*}

\section{Sample perfect output and errors}
\label{sec:app_error_analysis}

Figure \ref{fig:error_analysis} shows three user queries along with their generated workflows. The first one is a complicated workflow where the LLM is able to follow exactly the structure described in the user query, and is able to use the steps that the user expected. In this case, the retriever suggests only the step \texttt{post\_incident\_details}, as the rest are considered common steps.

In the second example, the retriever fails to suggest the \texttt{send\_slack\_message} step. The resulting workflow is not entirely wrong but it is of lesser quality as the LLM uses the common step \texttt{send\_notification}, which is not what the user expected.

In the last example, the LLM shows that it does not sufficiently understand the semantics of the task. The word \textit{Try} in the user query should have made it use the \texttt{TRY} and \texttt{CATCH} flow logic, but the LLM seems to ignore this word, resulting in a workflow that does not reflect what the user asked for.

\end{document}